\documentclass[11pt]{article}

\usepackage[final]{acl}

\usepackage{times}
\usepackage{latexsym}

\usepackage[T1]{fontenc}

\usepackage[utf8]{inputenc}

\usepackage{microtype}

\usepackage{inconsolata}

\usepackage{graphicx}

\usepackage{enumitem}

\usepackage{amssymb}
\usepackage{amsthm}
\usepackage{amsmath}
\newtheorem{definition}{Definition}

\definecolor{darkgreen}{RGB}{0,120,0}

\usepackage{booktabs}
\usepackage{tabularx}
\usepackage{multirow}
\usepackage[table]{xcolor}
\usepackage{array}
\usepackage{makecell}
\newcolumntype{Y}{>{\centering\arraybackslash}X}

\title{LaMoC: Loss-Aware Modular Compression for LLMs}

\author{
  Mohanad Odema\thanks{Corresponding author: \texttt{mohanad.odema@lge.com}} \\
  LG Electronics USA
  \And
  Jacob Song \\
  LG Electronics USA
}

\begin{document}
\maketitle
\begin{abstract}

Modular compression has enabled considerable parameter reduction in LLMs while preserving strong language understanding and downstream task accuracy. However, existing joint modular compression methods primarily rely on activation statistics, leaving loss-sensitivity information and its module-level characterization underexplored. We investigate addressing this gap with LaMoC, a loss-aware modular compression methodology that blends activation and Empirical Fisher statistics through gradient-error alignment. LaMoC improves joint compression by selecting compression statistics that better align local module reconstruction error with the downstream loss. Our contributions are three-fold: (1) We characterize the Empirical Fisher as a module-level loss-aware proxy that can be blended with the activation statistics required for compression. (2) We reformulate joint modular compression as a two-tiered optimization problem that minimizes module reconstruction error while tuning the activation and gradient information blending rate. (3) We implement an empirically driven methodology with statistical validation to solve the resulting compression problem. We evaluate LaMoC across four model families spanning eight models. On the 4–8B models, LaMoC achieves an average 2.5\% reduction in perplexity and a 1\% relative improvement in task accuracy over state-of-the-art modular compression methods.

\end{abstract}

\section{Introduction}

Parameter reduction is critical to enabling LLM efficiency in constrained deployment environments and at scale. Methods like pruning~\cite{ma2023llmpruner, ashkboos2024slicegpt, frantar2023sparsegpt, bai2024sparsellm} and low-rank approximation~\cite{wang2025svdllm, yuan2023asvd, wang2025dobi} have gained widespread adoption. Pruning removes weights, heads, or hidden dimensions based on a metric of importance in a structured or an unstructured manner, whereas low-rank approximation replaces the dense weight matrices with two low-rank factors that are more parameter- and compute-efficient. For example, Singular Value Decomposition (SVD) can approximate a single weight matrix as $W_{m\times n}\approx A_{m\times r}B_{r\times n} $ such that $r \ll \mathrm{min}(m, n)$, effectively reducing parameter count while providing optimal rank-$r$ truncation under Eckart-Young-Mirsky theorem \cite{eckart1936approximation, mirsky1960symmetric}. 

\begin{figure}[!t]
    \centering
    \includegraphics[width=\columnwidth]{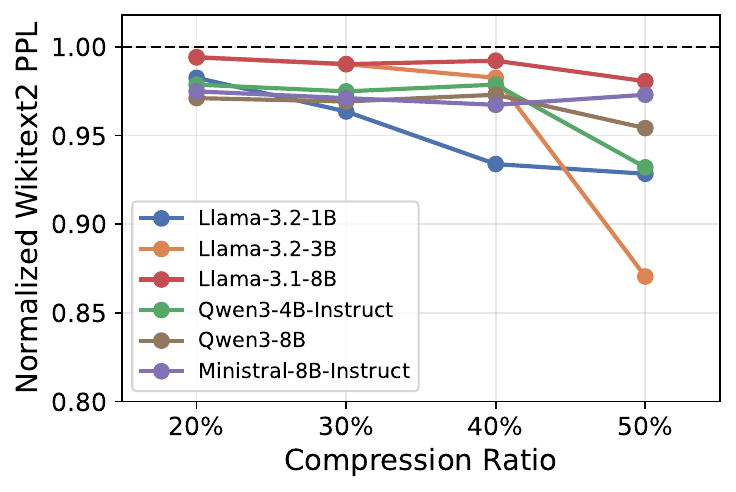}
    \vspace{-2ex}
    \caption{Normalized WikiText-2 perplexity for LaMoC relative to baseline MoDeGPT \cite{lin2025modegpt}.}
    \label{fig:motiv}
\end{figure}

Until recently, the considerable loss in knowledge capacity and task proficiency as a result of parameter reduction often necessitated a subsequent fine-tuning stage post compression. However, recent advances~\cite{lin2025modegpt, liu2025trainingfree, yin2025duogpt, chiang2026uniql} have substantially closed the post compression performance gap, reducing the reliance on expensive fine-tuning procedures, with several methods positioned as training-free approaches with an optional fine-tuning recovery stage. The effectiveness of these training-free methodologies can be attributed to these ideas: 

\begin{itemize}[leftmargin=*, itemsep=-2pt]
    \item \textbf{Activation Awareness.} 
    Modern compression methods increasingly move beyond weight-only criteria by using activation statistics to guide compression. This information can define the reconstruction target in low-rank approximation~\cite{wang2025svdllm,yuan2023asvd} or rank parameter importance for pruning~\cite{sun2024a}.

    \item \textbf{Loss Sensitivity Information. }Another thread follows the ideas of Optimal Brain Damage~\cite{lecun1989optimal, hassibi1993optimal}, where loss-derivative statistics are incorporated to assess parameter importance for compression, applicable in both pruning~\cite{ma2023llmpruner, yin2025duogpt} and low-rank reconstruction through loss sensitivity estimates~\cite{hsu2022language}.
    
    \item \textbf{Non-uniform Truncation. }Different layers and weight groups exhibit varying degrees of sensitivity to compression. Thus, recent approaches have opted to assign pruning/truncation ratios non-uniformly based on a measure of such sensitivity \cite{men2025shortgpt, yang2025let,li2025adasvd, wang2025svd, lin2025modegpt}. 
    
    \item \textbf{Modularity.} Rather than compressing each weight matrix independently, modular compression jointly compresses related weight groups, enabling module-level reconstruction objective and tailoring compression operators to each module's structural and functional properties~\cite{lin2025modegpt, wong2025a3, chiang2026uniql, koike2025latentllm}.
\end{itemize}

The combination of all four techniques remains underexplored -- specifically the incorporation of loss sensitivity information for modular compression. For instance, the state-of-the-art training-free compression method, MoDeGPT \cite{lin2025modegpt}, is a modular compression approach that leverages activation awareness and non-uniform truncation. In Figure \ref{fig:motiv}, we motivate how further consideration of loss sensitivity information on the module level can better preserve language modeling capabilities post compression, as seen by the decrease in perplexity compared to baseline MoDeGPT across different model architectures.  From here, we frame the following research questions: 

\begin{itemize}[leftmargin=*, itemsep=-2pt]

\item \textit{\textbf{RQ1:} Can loss-sensitivity information be systematically incorporated into modular compression to improve language modeling and downstream task performance?}

\item \textit{\textbf{RQ2:} How can module-level loss sensitivity be characterized and blended with the activation statistics required for joint compression?}

\item \textit{\textbf{RQ3}: 
How can joint modular compression be formulated as a loss-aware optimization problem to be solved in practical settings?}

\end{itemize}

Given these RQs, we summarize the key contributions of this work as follows:

\begin{itemize}[leftmargin=*, itemsep=-2pt]
\item We derive a module-level characterization to incorporate loss sensitivity information into modular compression using gradient-weighted second order activation statistics. 

\item We formulate the loss-aware modular compression objective as a two-tiered optimization problem minimizing the modular reconstruction loss while selecting a per-module blending rate between activation and gradient-weighted statistics.

\item We present LaMoC, an empirically-driven methodology with statistical validation to address the resulting two-tiered loss-aware modular compression problem leveraging fixed and adaptive strategies for controlling the degree of gradient information blending in a practical manner. 

\item We conduct experiments across 4 model families and 8 models showing that LaMoC improves SOTA modular compression pipelines by on average 2.5\% and 1\% in language modeling (perplexity) and downstream task accuracy, respectively.

\end{itemize}

\section{Background}

\subsection{Related Works}\label{sec:related}

\textbf{Parameter Reduction. }
Parameter reduction aims to improve LLM efficiency and memory footprint by removing or approximating less important parameters. This is commonly achieved through pruning~\cite{molchanov2019importance,xia2022structured,ashkboos2024slicegpt,frantar2023sparsegpt,sengupta2025you,ma2023llmpruner, men2025shortgpt} and low-rank approximation~\cite{wang2025svdllm,wang2025svd,wang2025dobi,yuan2023asvd,hu2026saessvd,hsu2022language}. Typically, pruning removes structured components such as layers, modules, channels, heads, or hidden dimensions, while low-rank approximation replaces dense weight matrices with compact factors. These methods define the basic parameter reduction operators. Recently, the application of these compression operators has extended beyond isolated matrices to coupled weight groups.  
 
\textbf{Modular Compression.}
Recent works~\cite{lin2025modegpt,chiang2026uniql,wong2025a3,koike2025latentllm} advance this direction through modular compression, where pruning or low-rank approximation is applied to structured groups of weight matrices according to their functional role and architectural properties. This modular view better reflects the underlying dynamics of coupled matrices whose outputs interact through non-linear operations and downstream transformations. As a result, modular compression introduces a broader module-level reconstruction objective rather than treating each matrix independently. Still, modular methods rely predominantly on activation-driven reconstruction criteria to guide the compression objective.

\textbf{Activation-Aware Compression.}
Instead of minimizing weight reconstruction error, modern parameter-reduction methods \cite{wang2025svdllm, wang2025svd, wang2025dobi, hu2026saessvd, lin2025modegpt, chiang2026uniql} target the reconstruction of output feature representations:
$\min \|X_l(W_l-\hat{W}_l)\|_F$, where $X_l$ is the input to layer $l$, and $W_l$ and $\hat{W}_l$ denote the original and compressed weights. Typically to achieve this in practice, calibration data is propagated through the model to collect second-order activation statistics (Gram autocorrelations), which are then used to project weights into the activation space or identify important dimensions. Recent works \cite{odema2026understanding} study the role of calibration data in residual error accumulation during compression and its effect on layer-sensitivity misalignment.
Still, these activation statistics do not necessarily capture how compression decisions affect the downstream loss.

\textbf{Loss-Aware Compression.}
A complementary line of work incorporates loss-derivative information to guide the compression decisions.
Early methods such as Optimal Brain Damage and Optimal Brain Surgeon~\cite{lecun1989optimal,hassibi1993optimal} use second-order derivative information to guide pruning decisions. The idea still persists in recent pruning~\cite{singh2020woodfisher, ma2023llmpruner, frantar2023sparsegpt} and low-rank approximation~\cite{hsu2022language} methods, where Hessian- or Fisher-based curvature proxies are used to estimate compression sensitivity and weight parameter importance.

\textbf{Motivation. }Despite progress in modular and loss-aware compression, their combination remains underexplored. Existing modular methods primarily rely on activation-driven reconstruction objectives, while loss-aware methods are often applied at the level of individual parameters or matrices. This motivates a study on loss-aware modular compression, how to incorporate loss-curvature statistics into module-wise compression objectives, and the potential gains from this approach. More details on the positioning of this work in relation to other loss-aware frameworks are in Appendix \ref{appdx:related_works}.

\begin{figure}[!t]
    \centering
    \includegraphics[width=0.9\columnwidth]{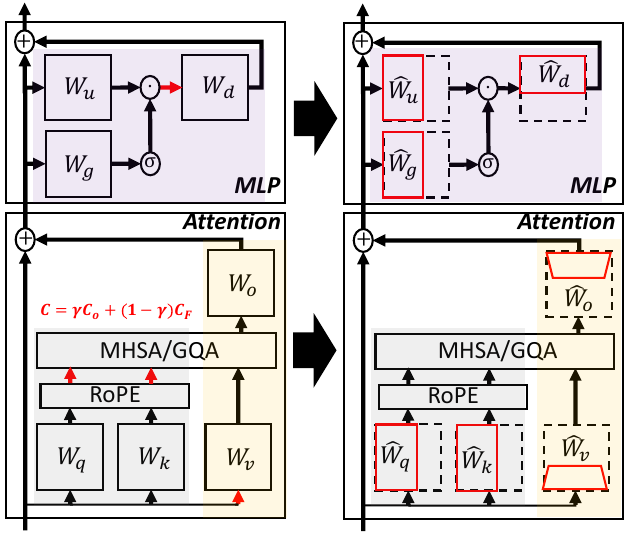}
    \caption{LLM Modular Compression. (\textit{left}) Colors demonstrate the different module groupings (QK, VO, MLP). Red arrows indicate the activation space in which the activation Gram $\mathcal{C}_0$ is collected and the empirical Fisher-weighted Gram $\mathcal{C}_F$ is projected into that same space from gradients computed at each module's output projection.
 The combined gradient-weighted Gram formulation is also shown in red. (\textit{right}) Weight matrices post Modular Compression.}
    \label{fig:joint}
\end{figure}

\subsection{Preliminaries on Joint Decomposition} \label{subsec:joint}

Figure \ref{fig:joint} illustrates the key module constructions for joint compression following recent SOTA works \cite{lin2025modegpt, chiang2026uniql, wong2025a3} detailed in the following.

\textbf{Notation. } Let $d_h$ denote the hidden dimension of the transformer model;  $d_{\mathrm{head}} = \frac{d_h}{n_{\mathrm{head}}}$ denote the head dimension given $n_{\mathrm{head}}$ attention heads\footnote{For simplicity, Notation shown is for multi-head attention}; $d_{\mathrm{int}}$ denote the intermediate dimension. The up and gate projection weight matrices are denoted by $W_u, W_g \in \mathbb{R}^{d_h\times d_{\mathrm{int}}}$, respectively; the down projection weight matrix is denoted by $W_d \in \mathbb{R}^{d_{\mathrm{int}}\times d_h}$; The query, key, and value matrices are denoted by $W_q, W_k, W_v \in \mathbb{R}^{d_h \times (n_{\mathrm{head}}\times d_{\mathrm{head}})}$; The output projection matrix $W_o \in \mathbb{R}^{(n_{\mathrm{head}}\times d_{\mathrm{head}})\times d_h}$. Broadly, activation Gram, matrices are denoted by $\mathcal{C}=X^\top X \in \mathbb{R}^{d\times d}$ where $X \in \mathbb{R}^{T\times d}$ indicate the input sequence of length $T$ and a dimension $d$ specified based on the where $X$ is captured. $\rho(\cdot)$ indicates Rotary Positional Embedding (RoPE). 

\textbf{MLP Module.} $W_u, W_g,$ and $W_d$ are jointly compressed through \textit{dimension pruning} of the shared intermediate dimension $d_{\mathrm{int}}$, reducing it to $r \ll d_{\mathrm{int}}$. In order to select which intermediate dimensions to prune, the following sequence is used: 1) Activation gram statistics $\mathcal{C}_{\mathrm{int}}=X_{\mathrm{int}}^\top X_{\mathrm{int}} \in \mathbb{R}^{d_{\mathrm{int}}\times d_{\mathrm{int}}}$ are collected prior to down projection using calibration data; 2) Per-dimension scores $\mathbf{s}\in \mathbb{R}^{d_{\mathrm{int}}}$ are computed based on criterion like ridge leverage scores $\mathbf{s}=\operatorname{diag}(\mathcal{C}_{\mathrm{int}}(\mathcal{C}_{\mathrm{int}} + \lambda I)^{-1}) \in \mathbb{R}^{d_{\mathrm{int}}}$; 3) The intermediate dimensions are sorted in a descending importance order based on $\mathbf{s}$; 4) A selection matrix from top-$r$ scores $S_k\in\mathbb{R}^{d_{\mathrm{int}}\times r}$ is constructed to be multiplied by the individual weight matrices to prune $d_{\mathrm{int}}$ providing $W_u S_k$, $W_g S_k \in \mathbb{R}^{d_h \times r}$ and $S_k^\top W_d \in \mathbb{R}^{r \times d_h}$.

\smallskip
\textbf{QK Module.} $W_q$ and $W_k$ are truncated along the head dimension $d_{\mathrm{head}}$. The key idea is to derive per-head scoring vectors, $\mathbf{s}_{\mathrm{head}}\in \mathbb{R}^{d_\mathrm{head}}$, before applying \textit{dimension pruning}. The scores are estimated based on channel importance evaluated at each attention head. Specifically, given an attention block input, $X \in \mathbb{R}^{T\times d_h}$, activation channel correlations are collected at each head $i$ through: 
\begin{itemize}[leftmargin=*, itemsep=-2pt]
\item \textbf{Query Correlations.} Let $X_q^i = \rho(XW_q^i)$ be the RoPE-transformed query activation for the $i^{th}$ attention head. Then, the query autocorrelation is $\mathcal{C}_q^{i}=(X_q^i)^\top X_q^i \in 
\mathbb{R}^{d_{\mathrm{head}}\times d_{\mathrm{head}}}$.
\item \textbf{Key Correlations.} Let $X_k^i = \rho(XW_k^i)$ be the RoPE-transformed key activation for the $i^{th}$ attention head. Then, the key autocorrelation is $\mathcal{C}_k^{i}=(X_k^i)^\top X_k^i$ $\in 
\mathbb{R}^{d_{\mathrm{head}}\times d_{\mathrm{head}}}$.
\item \textbf{QK Correlations.} Given $\mathcal{C}_k^{i}$ and $\mathcal{C}_q^{i}$, the final importance scores for each head $i$ are evaluated through a Hadamard product of the column norm vectors of their matrix square roots, $\mathbf{s}^i = ||(\mathcal{C}_q^{i})^{1/2}||_{\operatorname{col}} \odot ||(\mathcal{C}_k^{i})^{1/2}||_{\operatorname{col}} \in \mathbb{R}^{d_{\mathrm{head}}}$. 
\end{itemize}
Using $\mathbf{s}^i$, a per-head selection matrix $S_{qk}^i \in \mathbb{R}^{d_{\mathrm{head}}\times r}$ with $r\ll d_{\mathrm{head}}$ can be constructed to select the top-$r$ important channels: $\hat{W}_q^i = W_q^iS_{qk}^i$ and
$\hat{W}_k^i = W_k^iS_{qk}^i$, where
$\hat{W}_q^i,\hat{W}_k^i \in \mathbb{R}^{d_h \times r}$.

\smallskip
\textbf{VO Module.} $W_v$ and $W_o$ are jointly compressed through a sequence of SVD operations which render two low-rank matrices $\hat{W}_v \in \mathbb{R}^{d_h \times (n_{\mathrm{head}}\times r)}$ and $\hat{W}_o \in \mathbb{R}^{(n_{\mathrm{head}}\times r) \times d_h}$. To achieve this, activation Gram statistics are collected at the input to $W_v$, yielding $\mathcal{C}_{v}=X_{v}^\top X_v\in \mathbb{R}^{d_{h}\times d_{h}}$. For each head, two successive SVD operations are applied to approximate $\mathcal{C}_v^{1/2}W_v^iW_o^i$ as follows: (1) $(\mathcal{C}_v^{1/2}W_v^i)W_o^i = (U_v^i\Sigma_v^i(V_v^i)^\top)W_o^i$; (2) $U_v^i (\Sigma_v^i (V_v^i)^\top W_o^i) \approx U_v^i(U_r^i\Sigma_r^i (V_r^i)^\top)$, where the subscript $r$ denotes truncation to a rank $r$. The final low-rank matrices can then be defined as follows: $\hat{W}_v^i \leftarrow \mathcal{C}_v^{-1/2}U_v^iU_r^i$
and $\hat{W}_o^i \leftarrow \Sigma_r^i (V_r^i)^\top$.

\section{LaMoC: Loss-Aware Modular Compression for LLMs}

\begin{figure*}[!t]
    \centering
    \includegraphics[width=2\columnwidth]{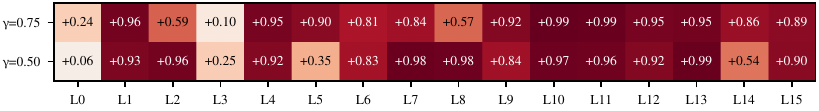}
    \caption{Pearson Correlation ($\uparrow$) between fixed reference first-order proxy and true $\Delta \mathcal{L}_{\mathrm{CE}}$ for Llama-3.2-1B layers.}
    \label{fig:mlp_corr}
\end{figure*}

This section introduces our method for loss-aware modular compression, which enforces gradient-error alignment during compression to incorporate loss sensitivity into the reconstruction objective. 

\textbf{Key Contribution and Overview. }The key contribution here lies in the formulation of loss-aware low-rank compression in the module activation space by blending gradient-weighted and canonical activation statistics, where a two-tiered compression objective is derived spanning module reconstruction and the gradient information blending rate. Based on empirical observations, we then propose fixed and adaptive blending strategies to control the blending rate in the module activation space according to an estimate on the expected change in cross entropy. 

\subsection{Gradient-Error Alignment}\label{subsec:gram_const}

Following the characterization of gram matrices for joint compression in Section \ref{subsec:joint}, we define a gradient-weighted effective Gram matrix. For readability, we omit the module superscript:

\begin{definition}
Let $\mathcal{C}_0=(X)^\top X$ be a canonical activation Gram matrix collected for an arbitrary module $l$. Define $\mathcal{C}_F$ as a Fisher-weighted activation gram, then an effective gradient-aligned activation Gram matrix is given as: 
\begin{equation}
    \mathcal{C}_{\mathrm{eff}}(\gamma) = \gamma \mathcal{C}_{0} + (1-\gamma) \mathcal{C}_F \label{eqn:eff_gram}
\end{equation}
where $\gamma \in [0,1]$ is a blending coefficient to control the degree of gradient information blending. 
\end{definition}

The red arrows in Figure~\ref{fig:joint} indicate for each module the activation space in which $\mathcal{C}_0$ and $\mathcal{C}_F$ are blended. This gradient-aligned Gram formulation can replace any activation Gram derived for any module from Section \ref{subsec:joint}. The canonical activation gram without any gradient information blending is obtained at $\gamma=1$. For $\gamma<1$, $\mathcal{C}_F$ skews the effective autocorrelation towards channels or energy directions more sensitive to the downstream loss. To provide a formal definition for $\mathcal{C}_F$, we first define a diagonal Fisher approximation as a per-channel importance vector:

\begin{definition}
    Let $\delta_n=\frac{\partial\mathcal{L}_{\mathrm{CE}}}{\partial x_n}\in \mathbb{R}^{d}$ be the gradient of cross entropy with respect to activation vector $x_n$ for the $n^{\mathrm{th}}$ calibration sample at a target compression module. Then, per-channel importance can be derived using the diagonal of empirical Fisher: 
    \begin{equation}
        \mathbf{f} = \operatorname{diag}(\mathbb{E}_{n}[\delta_n(\delta_n)^\top]), \qquad \mathbf{f} \in \mathbb{R}^{d}
    \end{equation}
which ignores cross-channel gradient correlations following the approximation in \cite{lecun1989optimal, hassibi1993optimal}.
\end{definition}

From the diagonal Fisher approximation we can construct a Fisher-weighted activation Gram: 
\begin{definition}
    Let $\mathbf{f}\in\mathbb{R}^{d}$ be the diagonal empirical Fisher vector characterizing the per-channel importance for a given module, then the Fisher-weighted activation Gram is defined as:
    \begin{equation}
    \mathcal{C}_F
    =
    D_f^{1/2}\mathcal{C}_0D_f^{1/2},
    \qquad
    D_f^{1/2}=\operatorname{diag}(\sqrt{\mathbf{f}})
    \end{equation}
\end{definition}
Lastly, we rescale $\mathcal{C}_F$ using a trace ratio to match the total energy of canonical $\mathcal{C}_0$ before blending: 
\begin{equation}
    \mathcal{C}_F \leftarrow \mathcal{C}_F \frac{\operatorname{trace}(\mathcal{C}_0)}{\operatorname{trace}(\mathcal{C}_F)} \label{eqn:trace}
\end{equation}
This ensures $\gamma$ blending between the canonical and Fisher-weighted statistics in Equation~\ref{eqn:eff_gram} is conducted on a matched energy scale. This concludes the gradient-aware effective Gram construction.

\subsection{Problem Formulation}

Given the loss-sensitive effective Gram matrix construction in
Section~\ref{subsec:gram_const}, the joint compression output reconstruction objective can be redefined.

\begin{definition}
Let $m$ be a target module to be jointly compressed, and let $\mathcal{C}_{\mathrm{eff}}(\gamma_m)$ denote the effective Gram defined in
Equation~\ref{eqn:eff_gram}. The module output reconstruction objective in a loss-aware joint compression process can be defined through the following two-tiered optimization formulation:
\begin{align}
    \hat{m}^\star &= \displaystyle{\arg\min_{\hat{m}}} ||({\hat{m}} - m)\mathcal{C}_{\mathrm{eff}}^{1/2}(\gamma^\star_{m})||_F^2 \\
    \displaystyle \gamma_{m}^\star &= \arg \min_{\gamma_{m}\in[0,1]} \Delta\mathcal{L}_{\mathrm{CE}} (\gamma_m) \label{eqn:gamma}
\end{align}
where $\hat{m}^\star$ represents the final compressed module conditioned on an optimal choice of $\gamma^\star_{m}$ that minimizes degradation in cross-entropy loss $\Delta \mathcal{L}_{\mathrm{CE}}(\gamma_{m})$ relative to canonical compression. 
\end{definition}

\begin{table*}[t]
\centering
\scriptsize
\setlength{\tabcolsep}{2.4pt}
\renewcommand{\arraystretch}{1.08}

\begin{tabularx}{\textwidth}{
l|
*{4}{>{\centering\arraybackslash}X}|
*{4}{>{\centering\arraybackslash}X}|
*{4}{>{\centering\arraybackslash}X}|
*{4}{>{\centering\arraybackslash}X}
}
\toprule
\multirow{2}{*}{\textbf{Method}}
& \multicolumn{4}{c|}{\textbf{Llama-3.1-8B}}
& \multicolumn{4}{c|}{\textbf{Ministral-8B-It-2410}}
& \multicolumn{4}{c|}{\textbf{Qwen3-4B-It-2507}}
& \multicolumn{4}{c}{\textbf{Qwen3 8B}} \\
\cmidrule(lr){2-5}
\cmidrule(lr){6-9}
\cmidrule(lr){10-13}
\cmidrule(lr){14-17}
& 20\% & 30\% & 40\% & 50\%
& 20\% & 30\% & 40\% & 50\%
& 20\% & 30\% & 40\% & 50\%
& 20\% & 30\% & 40\% & 50\% \\
\midrule

UniQL \cite{chiang2026uniql}
& 9.20 & 12.18 & 18.29 & 30.15
& 9.79 & 12.18 & 17.18 & 27.03
& 14.93 & 22.76 & 43.87 & 168.17
& 14.47 & 19.47 & 45.98 & 196.62 \\

MoDeGPT \cite{lin2025modegpt}
& 8.61 & 10.79 & 15.19 & 22.89
& 9.05 & 10.75 & 14.33 & 20.60
& 12.28 & 15.77 & 25.79 & 90.02
& 12.18 & 15.13 & 28.55 & 83.58 \\

\midrule

\rowcolor[gray]{0.95}
Fixed $\gamma=0.5$
& \textbf{8.54} & \textbf{10.67} & \textbf{15.07} & 22.54
& \textbf{8.83} & 10.46 & 13.89 & \textbf{20.05}
& 12.06 & 15.52 & 25.54 & 92.15
& 11.88 & 14.75 & 28.94 & 84.89 \\

\rowcolor[gray]{0.95}
Fixed $\gamma=0.75$
& 8.55 & 10.71 & \textbf{15.07} & \textbf{22.45}
& 8.86 & 10.50 & 13.94 & 20.16
& 12.13 & 15.67 & \textbf{25.24} & 89.32
& 12.06 & 14.90 & 28.27 & 82.60 \\

\midrule

\rowcolor[gray]{0.9}
Adaptive $\gamma^\star$
& 8.55 & 10.71 & \textbf{15.07} & \textbf{22.45}
& \textbf{8.83} & \textbf{10.44} & \textbf{13.86} & \textbf{20.05}
& \textbf{12.02} & \textbf{15.37} & \textbf{25.24} & \textbf{83.91}
& \textbf{11.83} & \textbf{14.67} & \textbf{27.78} & \textbf{79.75} \\

\bottomrule
\end{tabularx}

\caption{WikiText-2 perplexity ($\downarrow$) using 128 calibration samples from WikiText-2 at 2048 sequence length.}
\label{tab:compression_results}
\end{table*}

\begin{table}[t]
\centering
\scriptsize
\setlength{\tabcolsep}{3.0pt}
\renewcommand{\arraystretch}{1.08}

\begin{tabularx}{\columnwidth}{l|*{2}{>{\centering\arraybackslash}X}|*{2}{>{\centering\arraybackslash}X}}
\toprule
\multirow{2}{*}{\textbf{Method}}
& \multicolumn{2}{c|}{\textbf{Llama-3.2-3B}}
& \multicolumn{2}{c}{\makecell{\textbf{Llama-3.2-1B}}} \\
\cmidrule(lr){2-3} \cmidrule(lr){4-5}
& 20\% & 40\% & 20\% & 40\% \\
\midrule

UniQL \cite{chiang2026uniql}
& 14.93 & 47.44 & 27.03 & 184.70 \\

MoDeGPT \cite{lin2025modegpt}
& 10.88 & 22.32 & 17.52 & 44.21 \\

\midrule

\rowcolor[gray]{0.95}
Fixed $\gamma=0.5$
& 10.84 & 22.02 & 16.98 & 40.97 \\

\rowcolor[gray]{0.95}
Fixed $\gamma=0.75$
& 10.86 & 21.97 & 17.05 & 41.70 \\

\midrule

\rowcolor[gray]{0.9}
Adaptive $\gamma^\star$
& 10.81 & 21.93 & 17.21 & 41.29 \\

\bottomrule
\end{tabularx}

\caption{WikiText-2 PPL ($\downarrow$) for Llama-3.2-3B and Llama-3.2-1B models.}
\label{tab:compression_results_small}
\end{table}

\subsection{Method} \label{subsec:implement}

Solving the two-tiered optimization objective requires addressing the following challenges: 
\begin{itemize}[leftmargin=*, itemsep=-2pt]
    \item Navigating a combinatorial solution space comprising layers, modules, and $\gamma$ options
    \item Evaluation of $\Delta\mathcal{L}_{\mathrm{CE}}$ and the gradients for each candidate compression solution
\end{itemize}
We present our LaMoC method derived based on empirical analysis and statistical observations. 

\textbf{First-Order Loss Approximation.} \cite{lecun1989optimal} and \cite{hassibi1993optimal} used a Taylor Expansion to formulate the change in loss due to parameter reduction, generally defined as:
\begin{equation}
    \Delta \mathcal{L}_{\mathrm{CE}} = \delta^\top\Delta w + \frac{1}{2}\Delta w^\top \mathcal{H}\Delta w
\end{equation}
where $\delta$ indicates the gradient $\frac{\partial \mathcal{L}_{CE}}{\partial w}$ with respect to perturbed parameter; $\Delta w$ denotes parameter perturbation; and $\mathcal{H}$ the Hessian matrix. 
In our formulation, $\Delta \mathcal{L}_{CE}$ is only required to identify $\gamma^\star_{m}$ for a module $m$ as per Equation \ref{eqn:gamma}. Therefore, we can leverage a First-order approximation to evaluate the Expected $\Delta \mathcal{L}_{CE}$ post module $m$ compression without materializing the Hessian as follows: 
\begin{equation}
    \mathbb{E}[\Delta\mathcal{L}_{\mathrm{CE}}(\gamma_{m})] =  \frac{1}{N}\sum_{n=1}^N(\delta^{m}_{n})^\top X_n({\hat{m}_{\gamma_{m}}}-m) \label{eqn:CE}
\end{equation}
where $N$ is the number of input tokens; $X_n$ is the $n^{th}$ input at module $m$; $\delta_n^m$ is the gradient with respect to the $n^{th}$ module output; $\hat{m}_{\gamma_{m}}$ is compressed module version given $\gamma_m$. 

\textbf{Reference Gradients. }To avoid evaluating the $\delta_n^m$ with every sampled $\gamma$ in Equation \ref{eqn:CE}, we use the canonical solution gradients at $\gamma$=1 as the reference gradients to be used for all $\gamma$ solutions at a module $m$. Thus, the expected change in cross-entropy in Equation \ref{eqn:CE} becomes relative to the canonical solution whose gradients are only computed once, and reused for subsequent candidates. This enables faster evaluation without expensive backpropagation requirements for each $\gamma$ candidate. The reference gradients for proxy prediction are computed at the ($\gamma$=1.0) compression, and the activations are not updated to consider previous layer effects.

\textbf{Statistical Validation. }We validate the usage of fixed reference gradients in the first-order proxy for estimating $\Delta \mathcal{L}_{\mathrm{CE}}$ using a statistical analysis measuring Pearson correlation between the proxy estimates and the true cross-entropy changes. Figure \ref{fig:mlp_corr} illustrates the correlation at each layer's last module output for a Llama-3.2-1B model given $\gamma \in \{0.5, 0.75\}$. We observe the correlation strength differs per-layer as the proxy is dependent on the calibration settings and relies on a first-order approximation of reference gradients. Overall, we observe Pearson correlation reaches 0.91$\pm$0.28.

\textbf{$\gamma$ Selection Policy.} Denote $\mathbf{p}(\gamma_{m})=\mathbb{E}[\Delta\mathcal{L}_{\mathrm{CE}}(\gamma_{m})]$. Then, a heuristic $\gamma$ selection policy per module can be defined:
\begin{equation}
\gamma_{m}^\star =
\begin{cases}
\displaystyle \arg\min_{\gamma_{m} \in \Gamma_{m}} \mathbf{p}(\gamma_{m}),
& \text{if } \displaystyle \min_{\gamma_{m} \in \Gamma_{m}} \mathbf{p}(\gamma_{m}) < 0, \\[8pt]
1,
& \text{otherwise.}
\end{cases} \label{eqn:gamma_opt}
\end{equation}
where $\Gamma_{m}$ represents the set of viable discrete $\gamma$ candidates under exploration. The condition ensures that $\gamma$ is selected for the target module iff the proxy predicts a reduction in cross entropy loss relative to the canonical compression. Otherwise, canonical $\gamma_{m}=1$ is assigned to the module $m$. 

\textbf{End-to-end Compression.} Once $\gamma^\star_m$ is selected for each module, compression proceeds by applying the corresponding module-specific effective Gram and assembling the resulting modules into a gradient-aligned compressed model. We further validate the effects of compounded predictions on $\mathbf{p}(\gamma_{m})$ for all modules of Llama-3.2-1B after 15\% compression, and find that the overall Pearson correlation  still holds at 0.913.

\section{Experiments}

\begin{table*}[t]
\centering
\scriptsize
\setlength{\tabcolsep}{3.5pt}
\renewcommand{\arraystretch}{1.08}

\newcommand{\lightcell}[1]{\cellcolor[gray]{0.95}#1}
\newcommand{\darkcell}[1]{\cellcolor[gray]{0.9}#1}

\begin{tabular*}{\textwidth}{@{\extracolsep{\fill}} c|c|l|ccccccc}
\toprule
\textbf{Model} & \textbf{Compress.} & \textbf{Method}
& \makecell{\textbf{MMLU}\\\textbf{(5-shot)}}
& \textbf{ARC-e} & \textbf{ARC-c} & \textbf{PIQA}
& \textbf{WinoG.} & \textbf{HellaS.} & \textbf{Average} \\
\midrule

\multirow{7}{*}{\makecell{\textbf{Llama-3.1}\\\textbf{-8B}}}
& 0\% & Dense
& 65.60 & 81.57 & 53.75 & 80.25 & 74.35 & 78.89 & 72.40 \\
\cline{2-10}

& \multirow{3}{*}{20\%}
& MoDeGPT
& 55.83 & 69.74 & 41.04 & 71.98 & 71.74 & 67.45 & 62.96 \\
&
& \lightcell{Fixed $\gamma=0.75$}
& \lightcell{56.20} & \lightcell{71.17} & \lightcell{41.04} & \lightcell{72.03}
& \lightcell{71.03} & \lightcell{66.60} & \lightcell{63.01} \\
&
& \darkcell{Adaptive $\gamma^\star$}
& \darkcell{57.04} & \darkcell{70.96} & \darkcell{40.78} & \darkcell{71.87}
& \darkcell{70.56} & \darkcell{66.62} & \darkcell{62.97} \\
\cline{2-10}

& \multirow{3}{*}{40\%}
& MoDeGPT
& 41.61 & 48.53 & 29.78 & 64.31 & 63.54 & 48.89 & 49.44 \\
&
& \lightcell{Fixed $\gamma=0.75$}
& \lightcell{43.96} & \lightcell{50.63} & \lightcell{29.95} & \lightcell{63.44}
& \lightcell{64.09} & \lightcell{48.62} & \lightcell{50.12} \\
&
& \darkcell{Adaptive $\gamma^\star$}
& \darkcell{43.80} & \darkcell{50.80} & \darkcell{28.67} & \darkcell{62.73}
& \darkcell{62.75} & \darkcell{48.40} & \darkcell{49.52} \\

\midrule
\midrule

\multirow{7}{*}{\makecell{\textbf{Ministral}\\\textbf{-8B-It} \\ \textbf{-2410}}}
& 0\% & Dense
& 65.07 & 81.99 & 54.95 & 80.96 & 75.45 & 79.14 & 72.93 \\
\cline{2-10}

& \multirow{3}{*}{20\%}
& MoDeGPT
& 54.47 & 70.41 & 41.98 & 71.98 & 67.80 & 64.67 & 61.88 \\
&
& \lightcell{Fixed $\gamma=0.75$}
& \lightcell{55.88} & \lightcell{70.66} & \lightcell{41.98} & \lightcell{71.49}
& \lightcell{68.82} & \lightcell{64.62} & \lightcell{62.24} \\
&
& \darkcell{Adaptive $\gamma^\star$}
& \darkcell{55.95} & \darkcell{70.79} & \darkcell{42.24} & \darkcell{71.55}
& \darkcell{68.90} & \darkcell{64.72} & \darkcell{62.36} \\
\cline{2-10}

& \multirow{3}{*}{40\%}
& MoDeGPT
& 31.56 & 49.28 & 29.61 & 62.13 & 62.51 & 45.14 & 46.71 \\
&
& \lightcell{Fixed $\gamma=0.75$}
& \lightcell{31.44} & \lightcell{49.58} & \lightcell{29.10} & \lightcell{61.92}
& \lightcell{62.98} & \lightcell{44.82} & \lightcell{46.64} \\
&
& \darkcell{Adaptive $\gamma^\star$}
& \darkcell{31.94} & \darkcell{50.17} & \darkcell{29.27} & \darkcell{62.02}
& \darkcell{62.12} & \darkcell{44.98} & \darkcell{46.75} \\

\midrule
\midrule

\multirow{7}{*}{\makecell{\textbf{Qwen3-4B}\\\textbf{-It-2507}}}
& 0\% & Dense
& 72.60 & 83.21 & 58.70 & 76.01 & 67.88 & 69.09 & 71.25 \\
\cline{2-10}

& \multirow{3}{*}{20\%}
& MoDeGPT
& 46.26 & 70.92 & 46.33 & 72.09 & 62.75 & 62.77 & 60.19 \\
&
& \lightcell{Fixed $\gamma=0.75$}
& \lightcell{50.48} & \lightcell{69.91} & \lightcell{45.65} & \lightcell{72.25}
& \lightcell{63.77} & \lightcell{62.56} & \lightcell{60.77} \\
&
& \darkcell{Adaptive $\gamma^\star$}
& \darkcell{48.65} & \darkcell{73.57} & \darkcell{47.61} & \darkcell{71.55}
& \darkcell{63.77} & \darkcell{62.62} & \darkcell{61.29} \\
\cline{2-10}

& \multirow{3}{*}{40\%}
& MoDeGPT
& 24.65 & 46.21 & 29.27 & 61.37 & 55.96 & 42.69 & 43.36\\
&
& \lightcell{Fixed $\gamma=0.75$}
& \lightcell{24.90} & \lightcell{48.27} & \lightcell{30.55} & \lightcell{62.46}
& \lightcell{56.20} & \lightcell{42.89} & \lightcell{44.21} \\
&
& \darkcell{Adaptive $\gamma^\star$}
& \darkcell{24.80} & \darkcell{48.86} & \darkcell{29.35} & \darkcell{62.46}
& \darkcell{55.88} & \darkcell{42.87} & \darkcell{44.04} \\

\bottomrule
\end{tabular*}

\caption{Task Accuracy at 20\% and 40\% training-free compression using 128 samples from WikiText-2.}
\label{tab:compressed_llama_qwen_ministral}
\end{table*}

\textbf{Implementation. }Unless otherwise stated, we build our loss-aware modular compression on top of MoDeGPT with groupings of MLP, QK, and VO following Section \ref{subsec:joint} supplemented by LaMoC implementation in Section ~\ref{subsec:implement}. We follow MoDeGPT experimental setup with 128 calibration samples from WikiText-2 at sequence length 2048. 

\textbf{Selection Strategy. }We fix $\gamma_l \in \{0.5, 0.75\} \forall l\in \{1,\cdots,L\}$ and adopt two strategies for $\gamma$ selection: (1) \textit{Fixed}; where the same $\gamma$ value is assigned within each module in every layer. (2) \textit{Adaptive}; where each module selects $\gamma^\star$ value from the candidate set which achieves the minimum expected cross entropy loss following Equation \ref{eqn:gamma_opt}. 

\textbf{Baselines.} Our key baselines are MoDeGPT \cite{lin2025modegpt} and UniQL \cite{chiang2026uniql}. Both works follow in principle the joint decomposition flow introduced in Section \ref{subsec:joint} with some differences as UniQL is more system-oriented and relaxed some of MoDeGPT principled methodology. We focus on MoDeGPT and UniQL as they represent training-free methods shown to have outperformed strong pruning baselines (SliceGPT \cite{ashkboos2024slicegpt}, LLM Surgeon\cite{ouderaa2024the}, ShortGPT\cite{men2025shortgpt}).

\textbf{Evaluation and Models. }We evaluate the efficacy of our approach using WikiText-2 perplexity \cite{merity2016pointer}, 5-shot MMLU, and Zero-shot accuracy from the Lm-eval harness \cite{eval-harness} -- PiQA, Arc-E, Winogrande (accuracy); Hellaswag and Arc-C (normalized accuracy). We focus on the training-free compression performance following \cite{lin2025modegpt}, and target 1B-8B parameter models across the Llama-3 \cite{grattafiori2024llama}, Qwen3~\cite{yang2025qwen3}, and Mistral~\cite{ministral8b2410} families. We use base (Llama-3.1-8B, Llama-3.2-3B, Llama-3.2-1B) and instruct (Ministral-8B-Instruct-2410, Qwen3-4B-Instruct-2507\footnote{We use Qwen3-4B-Instruct-2507, Qwen3-4B-Inst., Qwen3-4B-It, and Qwen3-4B interchangeably in this work}, Qwen3-8B) models. Experiments are conducted on two NVIDIA Ada RTX 6000 with 48 GBs of memory each.

\subsection{Language Modeling Performance }

Table \ref{tab:compression_results} compares perplexity (PPL) for LaMoC against the aforementioned joint compression baselines given compression rates of 20\%, 30\%, 40\%, and 50\%. We draw the following observations: 

\textbf{Overall Performance. }Across all models and compression rates, LaMoC with gradient-error alignment improves the language modeling for all models across the different compression settings. The relative reduction in PPL compared to MoDeGPT reaches an average of 2.46\% for a relative reduction range of 0.79\% -- 6.79\%. The average absolute perplexity reduction is at 0.8 with an absolute PPL reduction range of 0.07-6.11.

\textbf{Model Type Influence. }The choice of model family exhibits an effect on the degree of improvement. Taking the 8B tier as an example, the relative reduction in PPL of Llama-3.1-8B is at 1.16\% (0.19 abs) contrary to Ministral-8B-It at 2.82\% (0.39), suggesting how original training mechanics and task prioritization could have an impact.  

\textbf{Model Size Impact. }Smaller models are more likely to benefit from the gradient-alignment approach to recover loss. The Qwen3-4B-It-2507 achieves a better average relative reduction of 3.39\% with an average absolute PPL reduction of 1.83. We verify this observation on the smaller Llama-3.2-1B and Llama-3.2-3B in Table \ref{tab:compression_results_small}, where the average reduction can reach 3.20\% for a fair subset of the compression ratios.

\textbf{Compression Rate Impact. }Gradient alignment benefits are more observable at aggressive compression rates, which can be attributed to the increasing loss scale. The scale of absolute PPL reduction increases as the compression rate increases -- from 0.23 at 20\% to 2.73 at 50\%.

\textbf{Choice of $\gamma$ Strategy. }The adaptive strategy offers the strongest relative reduction in PPL (2.42\%) compared to the fixed options (1.42\% and 1.52\%), respectively. For each configuration compared to the best fixed strategy, the adaptive $\gamma$ strategy leads to PPL reduction reaching up to 5.4 points compared to the best fixed $\gamma$ option.

\subsection{Task Proficiency Performance}

Table \ref{tab:compressed_llama_qwen_ministral} showcases the downstream task performance for the 5-shot MMLU and 0-shot tasks listed beforehand. We compare the performance at 20\% and 40\% for the Llama-3.1-8B, Ministral-8B-It-2410, and Qwen3-4B-It-2507 in the following. 

\textbf{Overall Performance. }Both fixed and adaptive modes of LaMoC improve over the baseline MoDeGPT. The improvements reach on average across all models and compression rates an increase of +0.99\% (+0.53 pp) across all benchmarks, broken down to +3.89\% (+1.65 pp) on the 5-shot MMLU, and +0.71\% (+0.40 pp) for the 0-shot accuracy. 

\begin{table}[t]
\centering
\small
\begin{tabular}{lcccc}
\toprule
Config & W & T & L & Mean acc. improv.\\
\midrule
Fixed $\gamma=0.75$      & 21 & 2 & 13 & $+0.41 \pm 1.07$ pp \\
Adaptive $\gamma^{\star}$  & 20 & 0 & 16 & $+0.40 \pm 1.11$ pp \\
\textbf{Best}              & \textbf{26} & \textbf{1} & \textbf{9} & $\mathbf{+0.67 \pm 1.11}$ \textbf{pp} \\
\bottomrule
\end{tabular}
\caption{Win/tie/loss counts and mean task accuracy improvement based on the data in Table \ref{tab:compressed_llama_qwen_ministral}.}
\label{tab:wtl}
\end{table}

\textbf{Statistical Analysis. }In Table \ref{tab:wtl}, we show the Win/Tie/Loss stats and average task accuracy improvement compared to the baseline based on the task accuracy data for the 6 benchmarks from Table \ref{tab:compressed_llama_qwen_ministral}. We observe that both Fixed $\gamma=0.75$ and Adaptive $\gamma^\star$ implementations achieve improvements over the baseline, with the best strategy choice achieving $+0.67\pm1.11$ pp accuracy improvement.

\textbf{Model Effect. }The smaller Qwen3-4B-It-2507 experiences the largest improvement +1.89\% (+0.98) with MMLU reaching +6.30\% (+2.24 pp),  compared to larger models like Llama-3.1-8B which improves by +3.65\% (+1.78 pp). This suggests how model type and size affects the degree of benefiting from gradient loss alignment given the varying information packing density per parameter translating into different scale of loss effects.

\textbf{Compression Rate Effect.} We also observe the improvements for the 40\% compression rate reaches +1.14\% (+0.52 pp) compared to the 20\% rate achieving +0.89\% (+0.54 pp). This goes in line with the insight that the benefits from gradient-loss aligned compression rise as the loss increases.

\subsection{Ablation}

\begin{table}[t]
\centering
\footnotesize
\setlength{\tabcolsep}{3pt}
\renewcommand{\arraystretch}{1.05}
\begin{tabular}{l|cc|cc|cc}
\toprule
 & \multicolumn{2}{c|}{Ministral-8B}
 & \multicolumn{2}{c|}{Qwen3-4B} 
  & \multicolumn{2}{c}{Qwen3-8B}
  \\
\cmidrule(lr){2-3} \cmidrule(lr){4-5} \cmidrule(lr){6-7}
 & $\gamma{=}1$ & $\gamma^\star$
 & $\gamma{=}1$ & $\gamma^\star$
 & $\gamma{=}1$ & $\gamma^\star$ \\
\midrule
\textbf{PPL} $\downarrow$       &  12.16 & \textbf{11.95} & 16.75 & \textbf{16.08} & 15.70 & \textbf{15.10} \\
\textbf{0-shot Avg} $\uparrow$  &  69.18 & \textbf{69.43} & 64.44 & \textbf{64.94} & 65.78 & \textbf{66.53}  \\
\textbf{MMLU} $\uparrow$  & 58.80 & \textbf{59.17} & 53.60 & \textbf{56.13} & 59.46 & \textbf{60.20} \\
\bottomrule
\end{tabular}
\caption{Calibration data ablation via Alpaca \cite{alpaca} at 20\% compression on instruction models.}
\label{tab:calibration_ablation}
\end{table}

\begin{table}[t]
\centering
\footnotesize
\setlength{\tabcolsep}{4pt}
\renewcommand{\arraystretch}{1.08}

\begin{tabular}{l|cc|cc}
\toprule
\multirow{2}{*}{}
& \multicolumn{2}{c|}{\textbf{Fixed $\gamma=0.75$}}
& \multicolumn{2}{c}{\textbf{Adaptive $\gamma^\star$}} \\
\cmidrule(lr){2-3} \cmidrule(lr){4-5}
& \textbf{EF} & \textbf{GGN}
& \textbf{EF} & \textbf{GGN} \\
\midrule

\textbf{PPL $\downarrow$}
& 41.70 & 41.46
& 41.29 & 41.05 \\

\textbf{LM-Avg $\uparrow$}
& 41.75 & 41.74
& 42.12 & 41.54 \\

\bottomrule
\end{tabular}

\caption{Loss ablation at 40\% comp. for Llama-3.2-1B.}
\label{tab:hessian_ablation}
\end{table}

\begin{figure}[!t]
    \centering
    \includegraphics[width=\columnwidth]{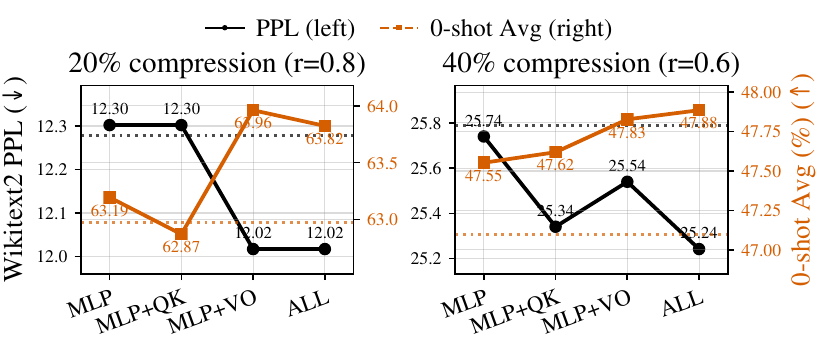}
    \vspace{-2ex}
    \caption{Per-module performance improvements across the different Qwen3-4B-It-2507 modules.}
    \label{fig:ablation}
\end{figure}

\textbf{Per-Module Contribution. }Figure \ref{fig:ablation} demonstrates performance contributions from LaMoC across the different modules from Qwen3-4B-It-2507 at 20\% and 40\% compression. We find that LaMoC application on MLP + VO offers the largest positive change in performance on both evaluation suites.

\textbf{Calibration Data.} In Table \ref{tab:calibration_ablation}, we ablate the choice of compression calibration dataset by using 128 samples from the Alpaca instruction dataset \cite{alpaca} on the instruction-tuned models from the aforementioned list of models. Two observations: (1) Adaptive $\gamma^\star$ wins on all 6 downstream tasks with average accuracy improvement of $+0.86\pm0.84$ pp compared to the baseline while being consistent across benchmarks – overall improvements reach 3.32\%, 0.75\%, and 2.12\% for perplexity, 0-shot average and MMLU, respectively. (2) Using Alpaca dataset for calibration yields a trade-off in language modeling and task accuracy, consistent with prior works' observations.

\textbf{Loss Signal Ablation.}
We ablate the loss signal used to bias the effective Gram, comparing Empirical
Fisher (EF) against generalized Gauss--Newton (GGN)~\cite{botev2017practical}.
Table~\ref{tab:hessian_ablation} reports results on Llama-3.2-1B under fixed $\gamma=0.75$ and adaptive
$\gamma^\star$. We observe GGN
trading off perplexity gains for LM-Avg reduction, seen in the adaptive
$\gamma^\star$ setting where PPL is reduced by 0.24, while LM-Avg is reduced by
0.58 points relative to EF.

\begin{table}[t]
\centering
\small
\begin{tabular}{llccc}
\toprule
Model & Ratio & Base & Trace $\checkmark$ & Trace $\times$ \\
\midrule
\multirow{2}{*}{Llama-3.2-1B}  & 0.8 & 17.52 & \textbf{17.21} & 17.52 \\
                           & 0.6 & 44.21 & \textbf{41.29} & 44.04 \\
\midrule
\multirow{2}{*}{Qwen3-4B-It}  & 0.8 & 12.28 & \textbf{12.02} & 12.28 \\
                           & 0.6 & 25.79 & \textbf{25.24} & 25.69 \\
\bottomrule
\end{tabular}
\caption{Trace-rescaling ablation: WikiText-2 PPL ($\downarrow$).}
\label{tab:trace_ppl}
\end{table}

\begin{table}[t]
\centering
\small
\begin{tabular}{llccc}
\toprule
Model & Ratio & Base & Trace $\checkmark$ & Trace $\times$ \\
\midrule
\multirow{2}{*}{Llama-3.2-1B}  & 0.8 & 49.94 & 49.15 & \textbf{49.95} \\
                           & 0.6 & 41.34 & \textbf{42.12} & 41.36 \\
\midrule
\multirow{2}{*}{Qwen3-4B-It}  & 0.8 & 62.97 & \textbf{63.82} & 62.98 \\
                           & 0.6 & 47.10 & \textbf{47.88} & 47.13 \\
\bottomrule
\end{tabular}
\caption{Trace-rescaling ablation: 0-shot average ($\uparrow$).}
\label{tab:trace_zeroshot}
\end{table}

\textbf{Trace Rescaling Ablation. }We ablate the effect of trace rescaling from Equation \ref{eqn:trace} by invoking compression with Adaptive $\gamma^\star$ with and without trace rescaling. Tables \ref{tab:trace_ppl} and \ref{tab:trace_zeroshot} show how the results change for WikiText-2 and average 0-shot accuracy for Llama-3.2-1B and Qwen3-4B-It. The results demonstrate the importance of trace rescaling to our method, where on average with trace rescaling, perplexity improvements rise from 0.19\% to 3.16\% while task accuracy gains rise from 0.02 to 0.40 pp.

\begin{table}[t]
\centering
\small
\begin{tabular}{llcc}
\toprule
Rate & Config & WikiT-2 PPL $\downarrow$ & 0-shot Avg $\uparrow$ \\
\midrule
\multirow{2}{*}{20\%} & UniQL & 6.73 & 69.74 \\
                      & \textbf{Adaptive $\gamma^{\star}$} & \textbf{6.68} & \textbf{71.16} \\
\midrule
\multirow{2}{*}{40\%} & UniQL & 10.58 & 54.00 \\
                      & \textbf{Adaptive $\gamma^{\star}$} & \textbf{10.42} & \textbf{55.25} \\
\bottomrule
\end{tabular}
\caption{Qwen2.5-32B evaluations post compression.}
\label{tab:qwen32b}
\end{table}

\textbf{Scalability. }We further assess the scalability of LaMoC by extending evaluations to Qwen2.5-32B \cite{qwen2.5} using 128 samples from WikiText-2 for calibration. For this tier of models, we use UniQL as our modular compression baseline, and perform our evaluations on an NVIDIA RTX PRO 6000 Blackwell with 96 GB of memory. Table \ref{tab:qwen32b} shows the results for 20\% and 40\% compression. We observe on average $+1.34$ pp improvements in the 0-shot average task accuracy. In Appendix \ref{appdx:exaone}, we show further scalability evaluations on EXAONE 4.5-33B \cite{choi2026exaone}.

\begin{table}[t]
\centering
\small
\begin{tabular}{llcc}
\toprule
Rate & Config & 
WikiT-2 PPL $\downarrow$ & 0-shot Avg $\uparrow$ \\
\midrule
\multirow{2}{*}{20\%} & Base & $12.29 \pm 0.01$ & $63.50 \pm 0.46$ \\
                      & Adaptive $\gamma^{\star}$ & $\mathbf{11.99 \pm 0.06}$ & $\mathbf{64.18 \pm 0.38}$ \\
\midrule
\multirow{2}{*}{40\%} & Base & $25.59 \pm 0.18$ & $46.76 \pm 0.61$ \\
                      & Adaptive $\gamma^{\star}$ & $\mathbf{24.96 \pm 0.25}$ & $\mathbf{47.60 \pm 0.35}$ \\
\bottomrule
\end{tabular}
\caption{Repeated Qwen3-4B-Instruct-2507 evaluations under 3 different seeds of calibration data.}
\label{tab:seed}
\end{table}

\textbf{Repeated Sampling. }In Table \ref{tab:seed}, we repeat our evaluation on the Qwen3-4B-Instruct-2507 at 20\% and 40\% using 3 different calibration sampling seeds to ensure consistency of the performance gains. We find that 0-shot accuracy improvement holds at $+0.76\pm0.26$ pp.

\subsection{Latency Performance Benchmarking}

\begin{table}[t]
\centering
\small
\begin{tabular}{l rr rr}
\toprule
 & \multicolumn{2}{c}{\textbf{Llama-3.1-8B}} & \multicolumn{2}{c}{\textbf{Qwen3-4B}} \\
\cmidrule(lr){2-3}\cmidrule(lr){4-5}
 & Dense & 20\% & Dense & 20\% \\
\midrule
\textbf{Prefill (ms)  }      & 341.1 & \textbf{278.9} & 185.0 & \textbf{165.2} \\
\textbf{Decode (ms/tok) }    & 21.91  & \textbf{20.2}  & 14.8  & \textbf{13.7}  \\
\textbf{Speedup(N=64) } & 1.0$\times$ & \textbf{1.1$\times$} & 1.0$\times$ & \textbf{1.1$\times$} \\
\bottomrule
\end{tabular}
\caption{End-to-end inference latency at 20\% modular compression via \texttt{torch.compile}. }
\label{tab:latency_compile_20pct}

\end{table}

We benchmark the end-to-end latency on the NVIDIA RTX Ada 6000 using \texttt{torch.compile}, setting batch size of 1 and 2048 sequence length. For prefill, we take the median of 20 runs, whereas for decode, we use greedy decode and take the median time per output token for 64 decode tokens.  In Table \ref{tab:latency_compile_20pct}, we show the results of the benchmarking at 20\% compression for the Llama-3.1-8B and the Qwen3-4B-Inst. On average, both models achieve 1.1$\times$ speedup compared to their dense baselines.

\section{Discussion}

Below are our key insights captured from gradient-loss information inclusion in modular compression.

\textbf{Performance gains stack. }We motivated this work by discussing the four key ideas in SOTA training-free compression frameworks: Activation-awareness, loss curvature information, non-uniform truncation, and modularity. This work demonstrated that the stacking of the four techniques can unlock further performance gains. 

\textbf{Generality.} LaMoC's proposal in modifying the activation statistics themselves makes it complementary to the techniques employed by any compression framework. LaMoC can scale to any framework that relies on activation statistics.  

\textbf{Future Directions. }LaMoC is based on empirical analysis and statistical observations, which limits our exploration to a small subset of candidate $\gamma$ values and a heuristic $\gamma$ selection method. Deriving an analytical method to solve for optimal $\gamma$ per each module can be investigated in future.   

\section{Conclusion}

We provided a characterization for loss-aware modular compression and presented LaMoC, a gradient-error aligned methodology for modular LLM compression. LaMoC injects loss sensitivity through gradient-weighted Gram matrices to address the two-tiered problem formulation over module reconstruction and gradient information blending rate. Our experiments have shown LaMoC improving over state-of-the-art modular compression methods across benchmarks and compression rates.

\section*{Limitations}

Our proposed methodology is derived based on empirical observations on the effects of incorporating gradient information onto the joint compression objective. Deriving a theoretically-proven analytical solution for $\gamma$ remains important for future exploration due to its potential for unlocking further performance gains and improving compression speed.
Our method has been tested and evaluated for models ranging from 1B to 33B in model size. Still, broader experimental validation at or beyond the $\geq30B$ parameter tier of models remains needed. Furthermore, evaluating the effectiveness of this methodology on emerging model architectures (Hybrid, State-space models, linear attention, Mixture of Experts) and tasks (long context reasoning, agentic tool calling) is a focus for future investigations.

\bibliography{custom}

\break

\appendix

\section{AI Usage}
\label{sec:ai}

The original ideas, methodology design, and experimental planning in the paper are from the authors. We use AI assistants in coding, writing assistance, formatting, and experimental design refinement.

\begin{table*}[t]
\centering
\small
\setlength{\tabcolsep}{6pt}
\renewcommand{\arraystretch}{1.15}
\newcolumntype{Y}[1]{>{\raggedright\arraybackslash\hsize=#1\hsize}X}
\begin{tabularx}{\textwidth}{l Y{0.80} Y{1.20} Y{1.00}}
\toprule
\textbf{Method} & \textbf{Loss approximation} & \textbf{Key contribution} & \textbf{Relation to LaMoC} \\
\midrule
OBD~\cite{lecun1989optimal} & Second-order & Introduces diagonal-Hessian saliency for loss-aware weight pruning & Establishes per-weight loss-aware compression \\
\addlinespace
OBS~\cite{hassibi1993optimal} & Second-order & Introduces inverse-Hessian pruning and weight loss compensation & Models weight interactions for pruning compensation \\
\addlinespace
WoodFisher~\cite{singh2020woodfisher} & Second-order; optional first-order term & Makes OBS-style pruning practical for larger networks by approximating the inverse empirical Fisher & Makes second-order pruning scalable \\
\addlinespace
LLM Surgeon~\cite{ouderaa2024the} & Second-order; optional first-order correction & Extends OBS-style pruning to LLMs using layer-wise activation and gradient statistics & Weights or dimensions pruning; LaMoC operates on module level \\
\addlinespace
LLM-Pruner~\cite{ma2023llmpruner} & First- and approximate second-order importance & Finds connected structures to be removed together, ranked by loss effect & Prunes module structures; LaMoC performs module-level low-rank approximation \\
\addlinespace
FWSVD~\cite{hsu2022language} & Diagonal empirical-Fisher-weighted reconstruction & Introduces row-wise Fisher-weighted low-rank approximation & Fixed row-wise Fisher weighting in weight space vs.\ adaptive activation--loss blending in module activation space \\
\bottomrule
\end{tabularx}
\caption{Positioning of LaMoC in relation to loss-aware compression works.}
\label{tab:related}
\end{table*}

\section{Detailed Related Works Positioning}\label{appdx:related_works}

We further elaborate on the relation of this work to existing related works mentioned in Section \ref{sec:related}. 

\subsection{Comparison to loss-aware approaches}

Table \ref{tab:related} provides a comprehensive comparison comparing LaMoC against existing works that consider Hessian/Fisher/gradient-based approaches to compression or pruning. 
LaMoC's key contribution lies in the formulation of loss-aware low-rank compression in the module activation space to blend gradient-weighted and canonical activation statistics. From there, LaMoC proposes heuristic-based fixed and adaptive strategies to control the blending degree, where the adaptive strategy uses gradient-error alignment proxy to predict the loss effect of each candidate reconstruction. 

\begin{table}[t]
\centering
\small
\begin{tabular}{lc}
\toprule
Method & Avg. 0-shot acc. $\uparrow$ \\
\midrule
ShortGPT \cite{men2025shortgpt} & 43.24 \\
SliceGPT \cite{ashkboos2024slicegpt} & 46.17 \\
SVD-LLM \cite{wang2025svdllm} & 54.20 \\
UniQL  \cite{chiang2026uniql}  & \textbf{66.56} \\
\bottomrule
\end{tabular}
\caption{Comparison against single-matrix compression and pruning  from UniQL~\cite{chiang2026uniql}}
\label{tab:joint_v_single}
\end{table}

\subsection{Modular compression landscape}

The majority of our experiments implement LaMoC on top of MoDeGPT \cite{lin2025modegpt} or UniQL \cite{chiang2026uniql} owing to them being state-of-the-art in modular compression. These works outperform conventional single layer low-rank approximation or pruning approaches. For instance, Table \ref{tab:joint_v_single} is from UniQL showing its superiority in compressing Llama-3.1-8B at 25\% rate compared to existing pruning or single matrix low-rank compression approaches, where UniQL maintains better average 0-shot accuracy post compression without any finetuning requirement. 

Additional frameworks that belong to the same MoDeGPT-style modular compression family include A3 \cite{wong2025a3} and LatentLLM \cite{koike2025latentllm}. UniQL was prioritized in this work based on the reported results of related works. The closest work was A3, where both UniQL and A3 shared a reporting on 5-shot MMLU for Llama-3.1-8B where UniQL achieved 60.2\% at 15\% compression compared to A3 which achieved 59.22\% at 10\% compression.

\section{Additional Experiments}

\begin{table}[t]
\centering
\small
\begin{tabular}{lcc}
\toprule
Config & WikiT-2 PPL $\downarrow$ & 0-shot Avg $\uparrow$ \\
\midrule
UniQL ($\gamma = 1.0$)         & 12.97          & 69.87 \\
Adaptive $\gamma^\star$ (Wiki)  & 11.81          & \textbf{70.37} \\
Adaptive $\gamma^\star$ (Mix)   & \textbf{11.62} & 70.14 \\
\bottomrule
\end{tabular}
\caption{Comparing EXAONE 4.5-33B results under 128 calibration samples from WikiT-2 and datamix (coding, Climbmix, reasoning) at 20\% compression}
\label{tab:diverse}
\end{table}

\begin{table*}[t]
\centering
\small
\begin{tabular}{lcccc}
\toprule
Config & \#Blends (max 16) & Oracle $\gamma$ agree & WikiT-2 PPL $\downarrow$ & 0-shot Avg $\uparrow$ \\
\midrule
Baseline ($\gamma=1$) & 0 & N/A & 19.65 & 54.98 \\
Adaptive $\gamma^{\star}$ ($\varepsilon=0$, $k=0$) & 14 & 7 & 17.31 & 54.60 \\
Adaptive $\gamma^{\star}$ ($\varepsilon=10^{-6}$, $k=3$) & 9 & 11 & 17.52 & 54.25 \\
Oracle & 9 & 16 & 17.53 & 54.68 \\
\bottomrule
\end{tabular}
\caption{Adaptive $\gamma^\star$ selection for Llama-3.2-1B MLP modules at $r=0.85$ compared to Oracle selection.}
\label{tab:gamma-oracle}
\end{table*}

\subsection{Calibration effects}
We evaluate the effects of changing the calibration dataset using the EXAONE 4.5-33B \cite{choi2026exaone} compared to the baseline UniQL. Specifically, we construct an alternative calibration data mix to collect the empirical-Fisher gradient grams, whereas the input covariance and layer ratio allocation remain calibrated through WikiText-2. The new calibration mix constitutes 128 samples distributed equally — 25\% Evol-CodeAlpaca (code), 25\% ClimbMix (English web text), 25\% Orca-Math English CoT (reasoning), and 25\% Korean (a seven-source in-house mix). The results at 20\% compression are shown in Table \ref{tab:diverse}, where we observe a slight capability trade-off between perplexity and downstream task proficiency. In either case, Adaptive $\gamma^\star$ remains superior to the baseline.

\begin{table}[t]
\centering
\small
\begin{tabular}{llcc}
\toprule
Model & Config & WikiT-2 $\downarrow$ & 0-shot Avg $\uparrow$ \\
\midrule
\multirow{2}{*}{Qwen3-4B} & MoDeGPT & 90.02 & 41.25 \\
                               & Adaptive $\gamma^{\star}$ & \textbf{83.91} & \textbf{41.76} \\
\midrule
\multirow{2}{*}{Ministral-8B}  & MoDeGPT & 20.60 & 44.76 \\
                               & Adaptive $\gamma^{\star}$ & \textbf{20.05} & \textbf{44.99} \\
\bottomrule
\end{tabular}
\caption{Instruction-tuned models 50\% compression with Adaptive $\gamma^\star$ compared to MoDeGPT ($\gamma=1.0$).}
\label{tab:aggressive_mid}
\end{table}

\begin{table}[t]
\centering
\small
\begin{tabular}{llcc}
\toprule
$r$ & Config & WikiT-2 PPL $\downarrow$ & 0-shot Avg $\uparrow$ \\
\midrule
\multirow{2}{*}{$0.8$} & UniQL ($\gamma=1.0$) & 12.97 & 69.87 \\
                       & Adaptive $\gamma^{\star}$ & \textbf{11.62} & \textbf{70.14} \\
\midrule
\multirow{2}{*}{$0.4$} & UniQL ($\gamma=1.0$) & 471.7 & 40.13 \\
                       & Adaptive $\gamma^{\star}$ & \textbf{356.0} & \textbf{42.59} \\
\midrule
\multirow{2}{*}{$0.3$} & UniQL ($\gamma=1.0$) & 1498.9 & 37.68 \\
                       & Adaptive $\gamma^{\star}$ & \textbf{1030.2} & \textbf{38.86} \\
\bottomrule
\end{tabular}
\caption{EXAONE 4.5-33B compression with Adaptive $\gamma^{\star}$ with focus on aggressive compression rates.}
\label{tab:aggressive}
\end{table}

\subsection{Aggressive Compression.}\label{appdx:exaone}

We evaluate aggressive compression across two tiers of models as follows:

\noindent
\textbf{Instruct Models (50\%).} We evaluate the Adaptive $\gamma^\star$ at 50\% compression for the Qwen3-4B-Instruct and Ministral-8B models in Table \ref{tab:aggressive_mid} compared to the baseline MoDeGPT \cite{lin2025modegpt}. We observe the Adaptive $\gamma^\star$ still outperforms the baseline in perplexity and 0-shot accuracy.

\noindent
\textbf{30B model tier (60\%-70\%).} We evaluate our Adaptive $\gamma^\star$ approach under 60\% and 70\% for EXAONE 4.5-33B \cite{choi2026exaone} compared to the baseline UniQL \cite{chiang2026uniql} modular compression approach. The results are shown in Table \ref{tab:aggressive}. We observe at both compression rates, the Adaptive $\gamma^\star$ improves both the WikiText-2 and 0-shot accuracy compared to the baseline UniQL.

\subsection{Adaptive $\gamma$ Selection} We study the viability of adaptive $\gamma$ as a heuristic-driven solution for the two-tiered problem formulation.  
From Section \ref{subsec:implement}, we first perform a detailed analysis on the selection proxy choices in Figure \ref{fig:mlp_corr} against an Oracle that is aware of the best $\gamma$ configuration per layer. 

\smallskip\noindent
\textbf{Generalized Adaptive $\gamma^\star$ formulation.} 
We define a more generalized form of the adaptive $\gamma^\star$ selection policy in Equation \ref{eqn:gamma_opt} with a confidence constrained blending condition incorporated as: 

\begin{equation}
\gamma_m^{\star}=
\begin{cases}
\arg\min\limits_{\gamma_m\in\Gamma_m}\mathbf{p}(\gamma_m),
& \begin{aligned}[t]
    &\text{if } \min\limits_{\gamma_m\in\Gamma_m}\mathbf{p}(\gamma_m)\\
    &< -\varepsilon - k\,\mathrm{SE}(\gamma_m),
  \end{aligned}\\
1, & \text{otherwise.}
\end{cases}
\label{eqn:gen_gamma_opt}
\end{equation}

where $\varepsilon$ and $k$ are constants; $\mathrm{SE}$ represents the standard error of the per-sequence proxy values across the calibration set, defined as: 
\begin{equation}
\mathrm{SE}(\gamma_m)
=
\frac{
\mathrm{sd}\!\left(\left\{p_i(\gamma_m)\right\}_{i=1}^{n}\right)
}{
\sqrt{n}
},
\end{equation}
where $n$ is the number of calibration data samples (here we use $n$=128 from Alpaca), $\mathrm{sd}$ is the standard deviation over the $n$ per-sequence proxy values $p_i(\gamma_m)$.

The generalized Equation \ref{eqn:gen_gamma_opt} applies the confidence threshold per candidate, where the candidate with the lowest predicted cross-entropy change among the valid options is assigned to $\gamma_m^{\star}$. This formulation allows blending only when the predicted cross-entropy reduction is (1) statistically significant, at least $k$ standard errors below zero, and (2) of non-negligible magnitude, exceeding $\varepsilon$.
 In other words, Equation \ref{eqn:gamma_opt} is a special case at $\varepsilon=0, k=0$.

\smallskip\noindent
\textbf{$\gamma$ choices analysis vs. Oracle}.
In Table \ref{tab:gamma-oracle}, we showcase the comparison of Adaptive $\gamma^\star$ selection approaches against the Oracle. 
From the table, we observe (1) the proxy can be more aggressive in picking $\gamma$ for gradient blending; (2) A more constrained implementation can better match the oracle choice, though not necessarily improve the performance; (3) Per-layer optimal $\gamma$ selection does not directly translate to pure downstream gains given other effects as compounded selection.

\begin{table}[t]
\centering
\small
\begin{tabular}{lccc}
\toprule
\textbf{EXAONE} & MLP & VO & QK \\
\midrule
Wiki ($r$=0.8)    & 51 $|$ 20 $|$ 31 & 42 $|$ 15 $|$ 27 & 42 $|$ 19 $|$ 23 \\
Mix ($r$=0.8) & 51 $|$ 28 $|$ 23 & 37 $|$ ~7 $|$ 30 & 43 $|$ 18 $|$ 25 \\
Mix ($r$=0.4) & 46 $|$ 26 $|$ 20 & 40 $|$ ~9 $|$ 31 & 48 $|$ 12 $|$ 36 \\
\bottomrule
\end{tabular}
\caption{$\gamma$ selection counts for EXAONE 4.5-33B (64 layers) at different calibration and compression. Entry format x|y|z reflects $\gamma<1.0$ $|$ $\gamma{=}0.75$ $|$ $\gamma{=}0.5$ choices.}
\label{tab:exaone_ngamma}
\end{table}

\begin{table}[t]
\centering
\small
\begin{tabular}{lccc}
\toprule
\textbf{Qwen3-4B-It} & MLP & VO & QK \\
\midrule
Wiki ($r$=0.8) & 22 $|$ 14 $|$ ~8 & 33 $|$ 12 $|$ 21 & 23 $|$ ~9 $|$ 14 \\
Wiki ($r$=0.6) & 32 $|$ 16 $|$ 16 & 33 $|$ 14 $|$ 19 & 24 $|$ ~8 $|$ 16 \\
\bottomrule
\end{tabular}
\caption{$\gamma$ selection counts for Qwen3-4B-Inst. (36 layers) at different calibration and compression. Entry format x|y|z reflects $\gamma<1.0$ $|$ $\gamma{=}0.75$ $|$ $\gamma{=}0.5$ choices.}
\label{tab:qwen4b_ngamma}
\end{table}

\textbf{Adaptive $\gamma$ Selection Statistics}. We report the $\gamma$ selection choices for EXAONE 4.5-33B and Qwen3-4B-Instruct where both outperform their respective baselines on the target tasks. For the EXAONE, we demonstrate the choices when using the WikiText-2 and datamix as the empirical Fisher calibration datasets. The results are displayed in Tables \ref{tab:exaone_ngamma} and \ref{tab:qwen4b_ngamma}, where we observe that the blending option offered through LaMoC is selected 70\%-77\% of the time across both tables.

\begin{table*}[!t]
\centering
\small
\renewcommand{\arraystretch}{1.2}
\begin{tabular}{llll}
\toprule
Statistic & MLP & VO & QK \\
\midrule
\makecell[l]{Gradient Gram $G$\\ {\footnotesize (per layer)}}
  & $[\mathrm{hidden},\,\mathrm{hidden}]$
  & $[\mathrm{hidden},\,\mathrm{hidden}]$
  & \makecell[l]{Q: $[n_q\!\cdot\!\mathrm{head}_{\mathrm{dim}},\; n_q\!\cdot\!\mathrm{head}_{\mathrm{dim}}]$\\
                 K: $[n_{\mathrm{kv}}\!\cdot\!\mathrm{head}_{\mathrm{dim}},\; n_{\mathrm{kv}}\!\cdot\!\mathrm{head}_{\mathrm{dim}}]$} \\
\midrule
\makecell[l]{Fisher diagonal $w$\\ {\footnotesize (VO: per KV group; QK: per head)}}
  & $[d_{\mathrm{int}}]$
  & $[\mathrm{head}_{\mathrm{dim}}]$
  & $[\mathrm{head}_{\mathrm{dim}}]$ \\
\midrule
\makecell[l]{Activation Gram $\mathcal{C}_0$,\\ Fisher-weighted Gram $\mathcal{C}_F$\\
             {\footnotesize (VO: per KV group; QK: per head)}}
  & $[d_{\mathrm{int}},\, d_{\mathrm{int}}]$
  & $[\mathrm{head}_{\mathrm{dim}},\, \mathrm{head}_{\mathrm{dim}}]$
  & $[\mathrm{head}_{\mathrm{dim}},\, \mathrm{head}_{\mathrm{dim}}]$ \\
\bottomrule
\end{tabular}
\caption{Tensor shapes per module. $G$ is collected once per layer at the module's output
projection; $w=\mathrm{diag}(W^{\top}G\,W)$ carries it back into the activation space
(for QK $w$ is simply the diagonal of $G$), where
$\mathcal{C}_F=\mathrm{diag}(\sqrt{w})\,\mathcal{C}_0\,\mathrm{diag}(\sqrt{w})$ shares
$\mathcal{C}_0$'s shape by construction. VO statistics are formed per KV group
($n_{\mathrm{kv}}$ per layer);
QK statistics are per head.}
\label{tab:tensor}
\end{table*}

\begin{table}[t]
\centering
\small
\begin{tabular}{lccc}
\toprule
Config & GSM8K$\uparrow$ & HumanEval$\uparrow$ & MT-Bench$\uparrow$ \\
\midrule
UniQL     & 38.2\% & 9.1\%  & 2.7\% \\
Adaptive $\gamma^{\star}$ & \textbf{39.6\%} & \textbf{14.6\%} & \textbf{3.7\%} \\
\bottomrule
\end{tabular}
\caption{Adaptive $\gamma^\star$ evaluations at $r=0.8$ compared to UniQL ($\gamma=1.0$) for free-form generation.} 
\label{tab:exaone_free}
\end{table}

\subsection{Free-form Generation Evaluation}
We further evaluate the EXAONE 4.5-33B on free-form generation benchmarks (GSM8K, HumanEval, MT-Bench) shown in Table \ref{tab:exaone_free} at 20\% compression compared to the UniQL baseline. We observe that the Adaptive $\gamma^\star$ outperforms the UniQL on the three benchmarks.

\begin{table}[t]
\centering
\small
\begin{tabular}{lcc}
\toprule
Config (wiki calibration) & PPL $\downarrow$ & 0-shot Avg $\uparrow$ \\
\midrule
Base ($\gamma=1.0$) & 12.28 & 62.97 \\
Adaptive $\gamma^{\star}$ (coarse) & 12.02 & \textbf{63.82} \\
Adaptive $\gamma^{\star}$ (fine)   & \textbf{11.83} & 63.69 \\
\bottomrule
\end{tabular}
\caption{Candidate $\gamma$ set granularity ablation, Qwen3-4B-Inst at 20\% compression, wiki calibration.}
\label{tab:grid_wiki}
\end{table}

\begin{table}[t]
\centering
\small
\begin{tabular}{lcc}
\toprule
Config (alpaca calibration) & PPL $\downarrow$ & 0-shot Avg $\uparrow$ \\
\midrule
Base ($\gamma=1.0$) & 16.75 & 64.41 \\
Adaptive $\gamma^{\star}$ (coarse) & \textbf{16.08} & \textbf{64.91} \\
Adaptive $\gamma^{\star}$ (fine)   & 17.69 & 64.51 \\
\bottomrule
\end{tabular}
\caption{Candidate $\gamma$ set granularity ablation, Qwen3-4B-Inst at 20\% compression, alpaca calibration.}
\label{tab:grid_alpaca}
\end{table}

\subsection{Candidate Selection Set Granularity}

We investigate the value from including additional $\gamma$ candidates into the $\gamma$ selection process. We expand the Adaptive $\gamma$ grid to include $\{0.125, 0.25, 0.375, 0.5, 0.625, 0.75, 0.875, 1.0 \}$. We repeat the experiment for WikiT-2 and Alpaca calibrations on Qwen3-4B-Inst. at 20\% compression in Tables \ref{tab:grid_wiki} and \ref{tab:grid_alpaca}. 
The key observations are: (1) Finer or coarser grids can both lead to accuracy gains against the baseline MoDeGPT; (2) The finer grid can be affected by predictor noise, coarser grid introduces generalization (see Table \ref{tab:grid_alpaca}).

\begin{table}[t]
\centering
\small
\begin{tabular}{lccc}
\toprule
Gradient & Relative cost & PPL $\downarrow$ & 0-shot Avg $\uparrow$ \\
\midrule
Baseline  & --             & 12.28 & 62.97 \\
Ref.\ gradient & $1\times$      & 12.30 & \textbf{63.19} \\
Exact gradient & $37.9\times$   & \textbf{12.28} & 63.12 \\
\bottomrule
\end{tabular}
\caption{Reference vs.\ exact gradients on Qwen3-4B-Instruct at $r=0.8$ with wiki
calibration (MLP modules). Relative cost is the number of backward passes normalized
to the reference-gradient setting.}
\label{tab:ref-grad}
\end{table}

\subsection{Reference Gradients Ablation}

We ablate the effect of using approximate reference gradients, computed once on the canonical ($\gamma=1.0$) compressed model and reused across all candidates, against exact gradients recomputed for each candidate, where the candidate's compressed weights are applied at its layer leaving all other layers in the canonical ($\gamma=1.0$) state. As shown in Table~\ref{tab:ref-grad}, reference gradients maintain performance close to that of exact gradients while requiring $37.9\times$ less backward-pass compute.

\subsection{Compression Times }

In Table \ref{tab:design_time}, we showcase the compression times taken by MoDeGPT and LaMoC when running their respective compression methods on the NVIDIA RTX Ada 6000 machine. LaMoC's additional compression time (2.5 h vs 3 h 10 min for Llama-3.1-8B) comes mainly from having to compute and store the gradients of the reference solution, and having to assess every candidate in the compression pipeline. The increase in time remains manageable with potential for further future optimization to target speeding up the operation.

\begin{table}[t]
\centering
\small
\begin{tabular}{l rr}
\toprule
& \textbf{Qwen3-4B} & \textbf{Llama-3.1-8B} \\
\midrule
MoDeGPT canonical       & 32 min     & $\sim$2.5 h    \\
\addlinespace
LaMoC: grads cache & +2 min     & +3 min         \\
LaMoC: candidates & +24 min  & +35 min        \\
\midrule
\textbf{Total LaMoC}                                   & \textbf{$\sim$58 min} & \textbf{$\sim$3 h 10 min} \\
\bottomrule
\end{tabular}
\caption{Wall time of LaMoC compression compared to canonical MoDeGPT at 20\% rate on NVIDIA RTX Ada 6000. LaMoC consumes additional time in: (1) Gradients caching from the canonical reference; (2) Evaluating the candidates in each module depending on the number of layers, $\gamma$ choices and number of modules. Analysis using 128 samples from WikiText-2.}

\label{tab:design_time}
\end{table}

\section{Tensor Shapes}
In Table~\ref{tab:tensor}, we provide the tensor shapes for the key data structures: the Gradient Gram $G$, the Fisher diagonal $w$, and the Activation and Fisher-weighted activation Grams ($\mathcal{C}_0$, $\mathcal{C}_F$) across MLP, VO, and QK modules. $\mathrm{hidden}$ is the model hidden dimension; $\mathrm{head}_{\mathrm{dim}}$ is the head dimension; $d_{\mathrm{int}}$ is the MLP intermediate dimension; $n_q$ and $n_{\mathrm{kv}}$ are the numbers of query and KV heads, respectively.

\end{document}